\documentclass{article}

\usepackage{microtype}
\usepackage{graphicx}
\usepackage{subfigure}
\usepackage{booktabs} 

\usepackage{hyperref}

\usepackage[accepted]{icml2024}

\usepackage{amsmath}
\usepackage{amssymb}
\usepackage{mathtools}
\usepackage{amsthm}

\usepackage[capitalize,noabbrev]{cleveref}

\theoremstyle{plain}

\theoremstyle{definition}

\theoremstyle{remark}

\usepackage[textsize=tiny]{todonotes}

\icmltitlerunning{Submission and Formatting Instructions for ICML 2024}

\begin{document}

\twocolumn[
\icmltitle{Many-Shot In-Context Learning for Molecular Inverse Design}




\begin{icmlauthorlist}
\icmlauthor{Saeed Moayedpour}{1}
\icmlauthor{Alejandro Corrochano-Navarro}{1}
\icmlauthor{Faryad Sahneh}{2}
\icmlauthor{Shahriar Noroozizadeh}{1}
\icmlauthor{Alexander Koetter}{3}
\icmlauthor{Jiří Vymětal}{7}
\icmlauthor{Lorenzo Kogler-Anele}{6}
\icmlauthor{Pablo Mas}{4}
\icmlauthor{Yasser Jangjou}{5}
\icmlauthor{Sizhen Li}{1}
\icmlauthor{Michael Bailey}{1}
\icmlauthor{Marc Bianciotto}{4}
\icmlauthor{Hans Matter}{3}
\icmlauthor{Christoph Grebner}{3}
\icmlauthor{Gerhard Hessler}{3}
\icmlauthor{Ziv Bar-Joseph}{1}
\icmlauthor{Sven Jager}{1}

\end{icmlauthorlist}

\icmlaffiliation{1}{Digital R\&D, Sanofi, Cambridge, MA, USA}
\icmlaffiliation{2}{Digital Data, Sanofi, Cambridge, MA, USA}
\icmlaffiliation{3}{Synthetic Molecular Design, Integrated Drug Discovery, Sanofi-Aventis Deutschland GmbH,
Industriepark Höchst, Building G838, 65926 Frankfurt am Main, Germany}
\icmlaffiliation{4}{Molecular Design Sciences, Integrated Drug Discovery, Sanofi, Vitry-sur-Seine, 94403, France}
\icmlaffiliation{5}{CMC Synthetics Platform, Sanofi, Cambridge, MA, USA}
\icmlaffiliation{6}{Digital R\&D, Sanofi, Toronto, Ontario, Canada}
\icmlaffiliation{7}{DataSentics, Brno, Czech Republic}

\icmlcorrespondingauthor{Ziv Bar-Joseph}{ziv.bar-joseph@sanofi.com}
\icmlcorrespondingauthor{Sven Jager}{sven.jager@sanofi.com}

\icmlkeywords{Large Language Models, Many-shot In-Context Learning, Molecular Design}

\vskip 0.3in
]



\printAffiliationsAndNotice{}  

\begin{abstract}
 Large Language Models (LLMs) have demonstrated great performance in few-shot In-Context Learning (ICL) for a variety of generative and discriminative chemical design tasks. The newly expanded context windows of LLMs can further improve ICL capabilities for molecular inverse design and lead optimization. To take full advantage of these capabilities we developed a new semi-supervised learning method that overcomes the lack of experimental data available for many-shot ICL. Our approach involves iterative inclusion of LLM generated molecules with high predicted performance, along with experimental data. We further integrated our method in a multi-modal LLM which allows for the interactive modification of generated molecular structures using text instructions. As we show, the new method greatly improves upon existing ICL methods for molecular design while being accessible and easy to use for scientists.
\end{abstract}

\section{Introduction}
Molecular design is fundamentally about crafting novel molecules that possess specific properties tailored to address particular scientific or industrial challenges \cite{Schneider2005}.
This process involves not only creating new molecular structures, but also refining and improving existing ones to more efficiently meet specific criteria. The refinement helps in converging a broad range of properties that the molecules need to satisfy, including affinity for particular biological targets, favorable pharmacokinetics and pharmacodynamics, or non-toxicological profiles, making them suitable for further development and eventual clinical trials. This is typically achieved through iterative cycles of synthesis, testing, and analysis, with each iteration aiming to bring candidates closer to the optimal properties. 

Every factor to be considered introduces additional layers of complexity and uncertainty, making the lead optimization and molecular design a resource-intensive and time-consuming task. Artificial intelligence (AI) emerges as a transformative tool capable of identifying and learning from complex patterns that can guide on the design of new molecules with optimized features. Furthermore, it can autonomously propose novel compounds that fulfill specific criteria, reducing the cycle time in iterative design and synthesis, and accelerating the development of new drugs \cite{Langevin2020, Loeffler2024Reinvent,Sauer2023reaction}.


%

Generative AI models are trained to understand and approximate the real distribution of the data they are fed with. Essentially, these models internally create a representation of the distribution such that the outcomes they produce adhere to the underlying rules governing the nature of the data \cite{https://doi.org/10.1002/wcms.1608}. In the context of chemistry, this 'data' refers to chemical space, which encompasses all possible chemical compounds and their configurations. Chemical space is vast and complex, containing an almost infinite variety of potential molecules with diverse physical and biological properties. Generative models can thus be used to propose novel molecules in an efficient and effective manner \cite{Meyers2021DeNovo}. 

Molecules can be represented using textual symbols corresponding to their structures. The most common description language for small molecules is SMILES (Simplified Molecular-Input Line-Entry System). Similar to other languages (including English and Amino Acids) several aspects of the SMILES language can be adequately captured by a LLM \cite{mirza2024superhuman}. Several examples of using SMILES based LLMs for supervised learning of small molecules have been presented \cite{ross2022largescale}.

Current initiatives to apply LLMs in the chemical space focus on two main areas, the prediction of chemistry-related downstream tasks \cite{blanchard2023LLM, Nascimento2023DoLLMs}, and the creation of self-operating agents with expertise in chemistry \cite{Boiko2023Emergent}. Recently, there have been a few studies on inverse design tasks with LLMs\cite{jablonka2024leveraging, liu2024drugllm, hocky2024connecting}. For example, \cite{wiest2023large} explores a text-based molecular design task. However, their prompt format does not use training data and only consists of a plain text description translated into a SMILES. 
ChemCrow \cite{Bran2023ChemCrow} is a recent agent-based method that introduces several innovations in AI-driven chemistry. By integrating 18 expert-designed tools, it enhances the performance of LLMs in various chemical tasks and reduces the risk of hallucinations. ChemCrow autonomously plans and executes chemical syntheses on a cloud-connected robotic platform, iteratively adapting and refining procedures without human intervention.

While their work shares the same purpose as ours in leveraging LLMs for chemical tasks, ChemCrow focuses on an iterative chain-of-thought process to autonomously plan and execute syntheses, emphasizing real-time adaptation and practical execution. In contrast, our molecular inverse design approach is geared towards providing structured data in a many-shot setup, enabling LLMs to learn the structure-activity relationship for designing molecules that satisfy multiple criteria. 



The significant increase in context size and reported findings suggests new possibilities for employing these techniques in complex domains such as molecular inverse design.In-context few shot learning (ICL) has emerged as a transformative strategy that allows LLMs to learn from a specific task at the time of inference without the need for specific fine-tuning \cite{agarwal2024manyshot}. Recent advancements have expanded the context windows available to LLMs, facilitating the usage of more examples for many shot learning. Unlike few-shot learning, many shot improve the model's ability to capture the underlying trends in the input data \cite{bertsch2024context}.

While many shot learning is an attractive training strategy, it faces the major challenge of obtaining additional experimental data, that is usually very expensive. To overcome this we developed a semi-supervised learning approach. Similar to the idea behind \cite{Mitchell1998}, the scarcity of labeled data forced us to develop different models trained on different views, i.e., set of features, to act as evaluators and predict the properties of untested molecules. By using different molecular descriptors for training these independent models we ensure robustness, as only those candidates whose predictions are consensual and meet success criteria are included in the pool for the next optimization round. Hence, we indirectly utilize the vast pool of newly-generated molecules to enhance our understanding and selection process without depending on experimental validation. However, it should be highlighted that these evaluators don't get retrained after every round, as they would potntially carry biases from the LLM in the generative process.




While fully automated design tool is the ultimate goal, humans still play a key role in the final design process. These are usually very experienced chemists with less computational background. For them, an online modification tool with deep understanding of structural and medicinal chemistry principles can greatly improve the overall outcome and the efficiency of the design process. We thus further extended the chemical design LLM we developed and combined it with a natural language LLM to provide fully interactive, instruction based, inverse molecule design tool.

\
\section{Methods}

\subsection{Datasets}
We used several datasets containing molecular activity data against 15 different pharmaceutically relevant proteins targets \cite{Sauer2023reaction}. Molecules were filtered by keeping only those compounds with specific activity data, i.e., IC\textsubscript{50}, K\textsubscript{i}, and posterior filtering by pCHEMBL values.

Next, molecules for each dataset were clustered and sorted based on activity using similarity of circular fingerprints. A property profiling of each cluster is conducted by including physicochemical and substructure properties, SA Score \cite{sascore} to estimate synthetic accessibility, or ligand efficiency among others. Top scoring cluster was selected and the 20 most active analogs from pCHEMBL (\textit{best20}) were chosen for experimental testing samples. In addition, the 50 most active compounds from the remaining clusters were used as lead structures for generative design (\textit{pool50}). An additional pool (\textit{allminus20}) is created containing all compounds belonging to the same target except for those being part \textit{best20}. 

Each pool contains the molecular structures represented as SMILES together with their corresponding activities, molecular weights, SA score, topological polar surface area (tPSA), and lipophilicity (logP).

\subsection{Molecular generation and evaluation details}

In order to perform many-shot in-context learning (ICL) experiments and design new molecules, we used the Claude 3 Sonnet model \cite{anthropic2024claude} with a 200k context window. Claude has been shown to perform well on math and reasoning benchmarks and provides a large context size that can accept up to 1 million tokens  \cite{anthropic2024claude}.  
In order to evaluate the generated molecules with our LLM framework, we used a combination of custom prediction models and scoring metrics available through MolScore\cite{thomas2023molscore}.
For activity prediction for each target, we trained several regression models using different types of input features. Section \ref{technical_details} provides details regarding the training and evaluation of the activity prediction models.
Several other properties, such as molecular weight, topological polar surface area (tPSA), logP, and synthesizability accessibility (SA) score, were also computed for the generated compounds.
Additionally, several intrinsic properties of the generated batches, such as internal diversity, as well as extrinsic properties, such as the Fréchet ChemNet Distance (FCD) of the generated batches from the lead molecule dataset, were calculated. To compare the performance of our framework with other state-of-the-art (SOTA) molecular design tools in the field, we performed lead optimization on the same datasets with REINVENT 4, which is a modern open-source generative AI framework for molecular design.

\subsection{Few-shot to many-shot ICL}

To assess the capability of LLMs for generating molecules with specific properties and to evaluate the performance gain by transitioning from few-shot to many-shot settings, we conducted several experiments. In the first experiment, we included 5 to 500 molecular SMILES strings and activity examples from the training data in the prompt to allow the LLMs to learn chemical patterns and substructures correlated with the activity. The provided examples in each setting were obtained by sorting the training data based on their activities and selecting the top performers. The model was then tasked with modifying a given lead molecule to generate new molecules with high activities and output them in a specific JSON format. Figure \ref{fig:icl_workflow} illustrates the structure of the input prompt in a multi-objective setting with several conditions on desired properties. 



\subsection{Iterative inclusion of self-generated data for ICL }

The training dataset contains a limited number of highly active molecules. In a many-shot setup with 100 training examples, the model generates novel and diverse molecules with high activities. However, when moving to larger sample size (500 molecules) the added shots, which make up the majority of examples, have low activity and the distribution of activities for generated molecules shifts toward lower activity regions.

To address this issue and utilize the full benefits of many-shot ICL, we explored an approach for the iterative inclusion of high-performing generated molecules and their predicted activities in the context, in addition to the experimental molecules and labels. 
For the predictive activity models, we trained several tree-based boosting models using three different types of features, namely circular fingerprints, RDKit descriptors, and Mol2Vec \cite{jaeger2018mol2vec} molecular features. 

Circular Fingerprints are generated using a hash function that transforms the local environment $E$ of each atom into a bit in a fixed-size vector. For each atom, the environment is defined by the type of atom and its neighbors up to a specified radius $r$:
 \begin{equation}
X_{cf}[i] = \bigoplus_{\text{atom } j \in \text{Atoms}(s)} \text{Hash}(\text{E}(j, \text{r}))
\end{equation}   

Here, $X_{cf}[i]$ is the $i$-th bit of the fingerprint vector, $\bigoplus$ denotes a bitwise OR operation across all atoms, and $\text{E}(j, \text{r})$ represents the substructure around atom $j$ within the given radius. $\text{Hash}$ is a function that maps each unique substructure to a bit position.

 RDKit computes a set of predefined descriptors for each molecule. These descriptors are functions of the molecular structure, such as molecular weight, logP, and the count of various types of bonds and atom types. The generation of a descriptor vector can be mathematically represented as:
\begin{equation}
   X_{rd} = \left[f_1(s), f_2(s), \ldots, f_n(s)\right]
\end{equation}

Where $f_k(s)$ is the $k$-th descriptor function applied to the SMILES string $s$, and $n$ is the total number of descriptors calculated by RDKit.

 Mol2Vec converts each molecule into a vector by first tokenizing the SMILES string into meaningful chemical fragments and then mapping each fragment to a vector in a pre-trained embedding space:
\begin{equation}
   X_{mv} = \frac{1}{m} \sum_{k=1}^{m} \text{Vec}(\text{Token}_k)
\end{equation}

In this equation, $m$ is the number of tokens in the SMILES string $s$, $\text{Token}_k$ is the $k$-th token, and $\text{Vec}(\text{Token}_k)$ is the vector representation of $\text{Token}_k$ from a pre-trained embedding model. The final feature vector $X_{mv}$ is the average of all token vectors, providing a dense representation of the molecule.

Each feature set $X_{cf}$, $X_{rd}$, and $X_{mv}$ is then used as input to separate gradient boosting models, which are trained to predict molecular activity $a'$ based on these distinct representations:

\begin{equation}
M(X) \rightarrow a'
\end{equation}

Where $X$ could be any of the feature vectors and $y$ is the predicted activity. The performance of the predictive models is shown in Figures \ref{fig:model_ensemble_nonw} and Figure \ref{fig:pairplot} provides pairwise correlation plots of predicted activities using these models.

\subsection{Semi-Supervised training for few shot learning}
To increase the number of high active samples we  designed an iterative process in which we included newly generated molecules whose predicted activity from all three activity prediction models was higher than a cutoff. As discussed above, while there is a small overlap in the input features used by the three methods their are also unique features used by each one. For the selected generated molecules, the average predicted activity was used as the final activity label in the context. We define $M$ as an LLM model that generates new SMILES strings based on the current context $S$ and corresponding activities $A$. The function $M$ operates as follows:

\begin{equation}
M(S, A) \rightarrow S'
\end{equation}

The initial context, $S_0$, consists of 500 experimental molecules represented by SMILES strings along with a list of their corresponding activities $A_0$:
\begin{align}\label{FCD}
S_0 &= \{s_{0,1}, s_{0,2}, \dots, s_{0,500}\}\\ 
A_0 &= \{a_{0,1}, a_{0,2}, \dots, a_{0,500}\}
\end{align}

During each iteration $i$, the context $S$ is expanded by including newly generated SMILES strings $S'$ with predicted activities $A'$ that exceed a cutoff $C$ value determined by the 80th percentile of activities from the current context:

\begin{align}
S_{i+1} &= S_i \cup \{s' \in S' \mid a' > C_i \wedge s' = M(S_i, A_i)\} \\
A_{i+1} &= A_i \cup \{a' \mid a' > C_i \wedge a' = M(S_i, A_i)\}
\end{align}




Using this approach, the context expands by LLM generated molecules with high predicted activity. We observed a improvement in the distribution of activities of generated molecules within a few iterations. The results are further presented and discussed in section \ref{iterative_results}.

\subsection{ICL for multi-objective molecular design}

To further verify the capabilities of LLMs for learning the relationships between chemical structures and specified properties, especially in multi-objective settings, we performed experiments where additional property conditions on molecular weight, synthetic accessibility (SA) score, tPSA, and logP were imposed in the context in addition to activity.

We designed two additional types of experiments for the multi-objective setup. In the first type, for a different property such as SA score, we set the requirement in the generation criteria section of the prompt, such as a low SA score (below 3, with lower values being better), without including any SA score labels in the context. In the second experiment, we included both the SA score requirement and the SA score values of the training examples. This experiment was designed to understand if LLMs are capable of learning within the context and identifying patterns between chemical structures and the property values or if they rely solely on pretrained information regarding different chemical properties when asked for a certain property condition in the requirement section of the prompt. Finally, we explored the performance of our workflow for molecular design by including up to 5 different criteria on activity, molecular weight, SA score, tPSA, and logP, and analyzed the properties, validity, novelty, and diversity of the generated molecules. The results are presented in section \ref{multiobjective_results}.

\subsection{ICL for property prediction}

We investigated the ability of SMILESs based LLMs to learn quantitative structure-activity relationships (QSAR) through in-context learning (ICL) for activity prediction tasks. To evaluate this, we performed cross-validated experiments where we compared the performance of LLMs for activity prediction tasks through many shot ICL with different CatBoost regression models. These models were trained using described input features, including circular fingerprints, RDKit descriptors, and Mol2vec features. To use LLMs for predicting activities, we provided the SMILES and activity examples from the training datasets and asked the model to predict the activity of a certain SMILES in the test datasets. We repeated this query for every individual SMILES in the test dataset, as well as all train/test datasets of the cross-validation folds. The results showcasing the capability of LLMs for QSAR and predicting activities from SMILES are discussed in Section \ref{res:property_prediction}.

\subsection{Interactive molecular design}

Our proposed ICL workflow can successfully generate novel candidate molecules with multiple target properties. However, the generated molecules are not always ideal. In many instances, a generated molecule with high predicted performance may be potentially appealing to an expert chemist, but could be significantly enhanced by a minor modification, particularly in terms of synthesizability. In such cases, directly modifying the SMILES string, especially for large complex molecules, can pose challenges. Following our observations of the profound chemical understanding of LLMs, we have developed an online interactive design module that allows for seamless modification of generated molecules using text instructions. This module can take the SMILES representations of molecules and the 2D visualisation of the molecule as inputs and perform specified modification on structural or physicochemical properties of the molecule. This design module can also be linked with customized databases containing desired molecular properties to expand its capability for improving that particular property through manyshot-ICL.

\section{Results and Discussion}

\begin{figure}[ht!]
    \centering
    \includegraphics[width=0.8\linewidth]{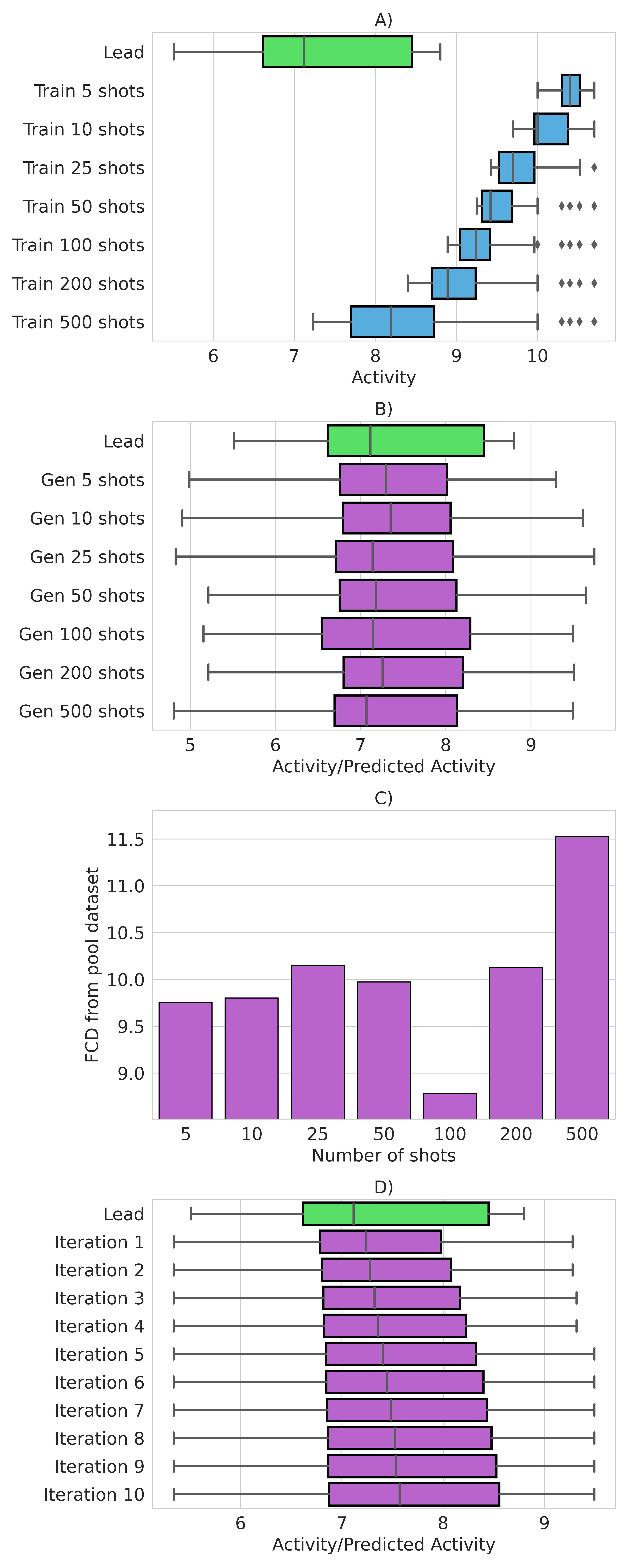}
    \caption{ A) Distribution of molecular activities against MMP8 protein target in the lead molecules dataset (Lead) and the subset of training datasets that include the top 5 to 500 highly active molecules.
B) Distribution of activities in lead molecules and predicted activities of generated molecules with 5 to 500 shots. C) FCD distance of generated molecules from the lead molecules in 5 to 500 shot ICL experiments. D) Distribution of activities in lead molecules and predicted activities of generated molecules in different iterations of including self-generated molecules and predicted activities in the context.
    }
    \label{fig:icl_results}
\end{figure}


\begin{figure*}[h!]
    \centering
    \includegraphics[width=0.8\textwidth]{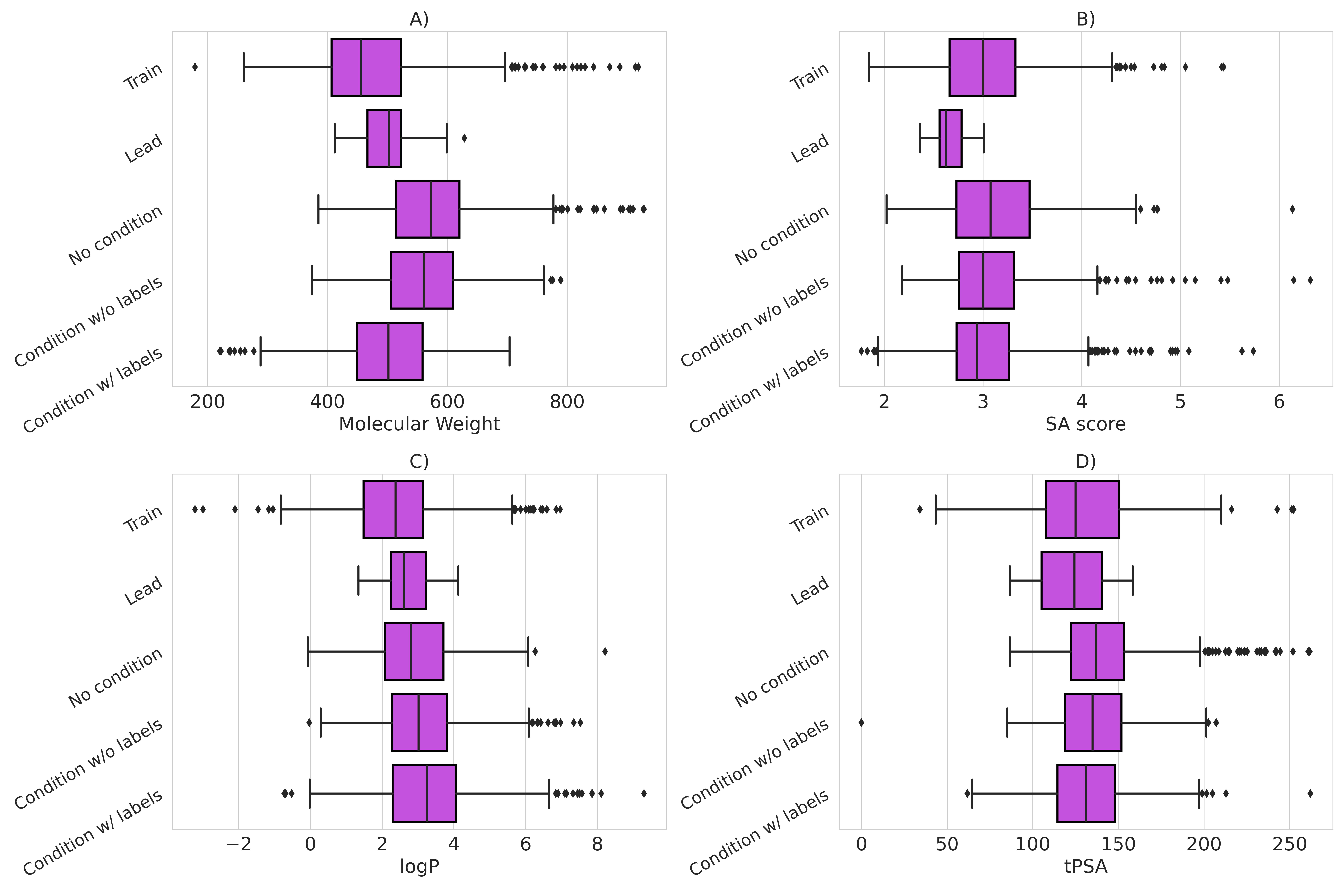}
    \caption{ Distribution of (a) molecular weight, (b) SA score, (c) logP, and (d) tPSA in the training dataset (Train), lead dataset (Lead), 500-shot ICL without any property criteria other than activity (No condition), 500-shot ICL with specified property condition without providing the property labels in the context (Condition w/o labels), and 500-shot ICL with specified property condition and provided the property labels in the context (Condition w/ labels).
    }
    \label{fig:multi_objective}
\end{figure*}

\subsection{Iterative many-shot ICL with self generated data for molecular design}\label{iterative_results}

Figure \ref{fig:icl_results}A compares the distribution of molecular activities against the MMP8 protein target in the lead molecules dataset (Lead) and the utilized subset of training datasets in the 5 to 500 shot ICL experiments, which include the top 5 to 500 highly active molecules. As can be seen, there is only a limited number of highly active molecules. This leads to performence decline when using more than the top 10 molecules (Figure \ref{fig:icl_results}B). The upper quartile of the distribution of generated molecules still increases with more molecules, but even the top ones deteriorate when including more than 100 shots (Figure \ref{fig:icl_results}B). In contrast, for the iterative design learning that we developed results continue to improve even after 10 iterations and 1125 molecules. In addition, the FCD distances of generated molecules from the pool of molecules significantly increase in the iterative procedure (Figure \ref{fig:icl_results}C),
demonstrating the generation of more novel candidates due to the model's access to a wider variety of chemical fragments in the context.

To utilize the full potential of many-shot ICL for generating novel molecules with high predicted activities, we performed iterative ICL experiments. In each iteration, in addition to 500 shots of experimental data, we added generated molecules with high predicted activity from the previous iteration. The selected generated molecules have a predicted activity above a cutoff (80th percentile of activities in the training data) by all three trained activity prediction models. We observed a shift in the distribution of activities of generated molecules toward the high activity region within a few iterations, as shown in Figure \ref{fig:icl_results}D.


\subsection{Multi-objective molecular design with many-shot ICL}\label{multiobjective_results}
\begin{figure*}[ht!]
\centering
\includegraphics[width=0.95\linewidth]{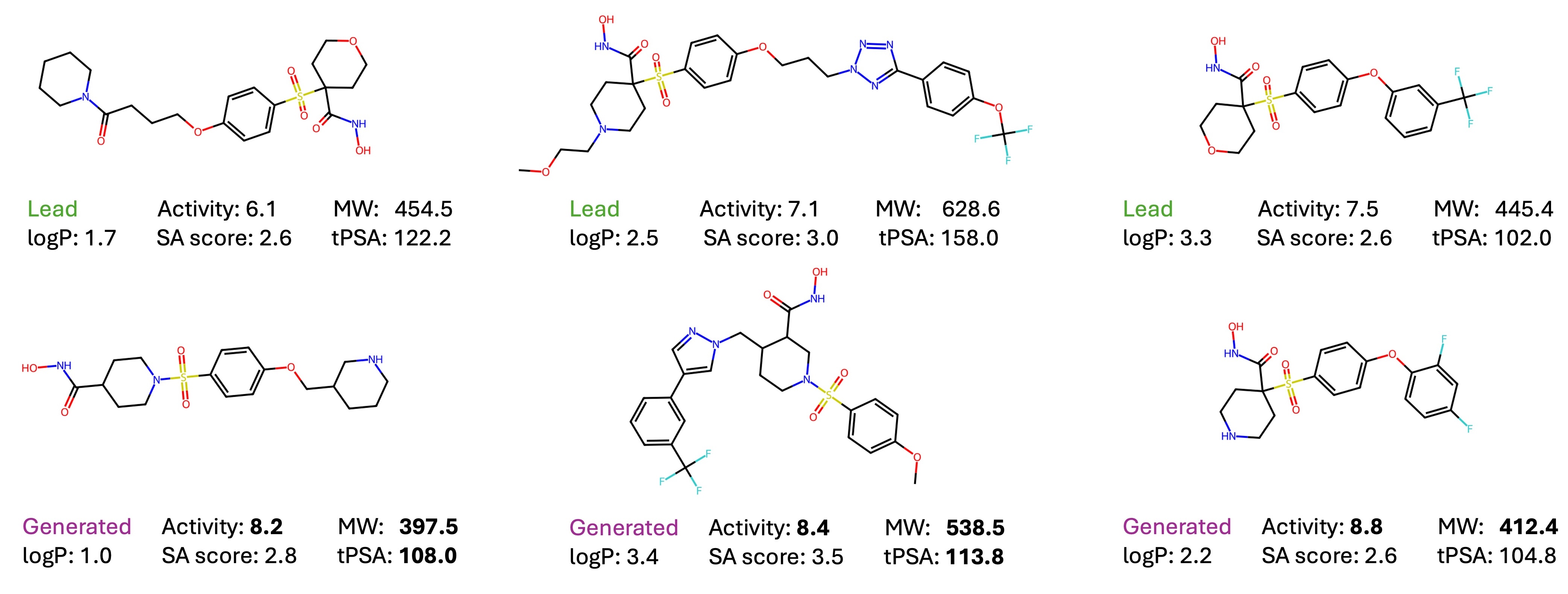} %
\caption{Example lead molecules and the resulting generated molecules along with their predicted properties. The improved properties are highlighted.}
\label{fig:lead_gen}
\end{figure*}

Figure \ref{fig:multi_objective} presents the distribution of different properties in the 500-shot ICL generation experiments where, in addition to molecular activities, additional property conditions and labels were introduced in the context. All generation experiments shown in Figure \ref{fig:multi_objective} include a condition on generating highly active molecules (activity above 10, with higher values being better) and also a secondary condition on molecular weight between 320-420 (Figure \ref{fig:multi_objective}A), low synthetic accessibility (SA) score (under 3, with lower values being better) (Figure \ref{fig:multi_objective}B), logP between 2 to 4 (Figure \ref{fig:multi_objective}C), and tPSA between 40-60 (Figure \ref{fig:multi_objective}D). The specified conditions on molecular weight, logP, and tPSA criteria were obtained from the work by Sauer et al.\cite{Sauer2023reaction}

We observed that the LLM can successfully perform ICL with additional property conditions and labels and generate new molecules with multiple target properties. Another notable observation was that providing an additional property condition in the prompt without providing any labels for the additional property led to the generation of molecules with properties closer to our desired range. This improvement is more significant in the case when property labels for additional property conditions are provided, which confirms the capability of LLMs for performing ICL, even though they appear to have gained impressive quantitative structure-property relationship (QSPR) knowledge throughout their pretraining. 

Finally, we conducted a 500-shot multi-target design experiment incorporating all five property conditions: activity, molecular weight, logP, synthetic accessibility (SA) score, and topological polar surface area (tPSA). By including these conditions in the prompt, our workflow successfully shifted the properties of the generated molecules towards the specified ranges. Figure \ref{fig:lead_gen} illustrates sample lead molecules and the resulting generated molecules. As shown, the model managed to generate molecules with higher predicted activities, lower desired molecular weights, lower desired tPSA values, and logP and SA scores within the acceptable ranges.

\subsection{Molecular property prediction with many-shot ICL}\label{res:property_prediction}

Our results demonstrate that Large Language Models (LLMs) can effectively learn the quantitative structure-activity relationships through ICL. The performance comparison of activity prediction with MMP8 protein target for LLMs and CatBoost models, which were trained with different input features, is presented in Table 1. Additionally, Figure 1 provides visual representation of the scatter plots of predicted activities obtained by different models against the experimental values of activity for the same protein target.Although LLMs did not surpass the performance metrics of the CatBoost models, their ability to learn structure activity relationships was evident. These results are in agreement with their capability for generating molecules with target properties through many-shot ICL, showing that LLM can leverage the learned structure-activity relationships to generate new highly active molecules.

\begin{table}[h!]
\centering
\begin{tabular}{||c c c c||} 
 \hline\hline
 Model & Input & $R^2$ & RMSE  \\  
 \hline
 LLMs & SMILES & 0.651 & 0.893  \\ 
 CatBoost & Circular Fingerprints & 0.784 & 0.710 \\
 CatBoost & RDKit descriptors & 0.777 & 0.723  \\
 CatBoost & Mol2vec features & 0.753 & 0.760  \\
 
  \hline
\end{tabular}
\caption{Regression Performance Metrics of Different Models for Activity Prediction for MMP8 protein target. The table provides the Average 10-fold Cross-Validated $R^2$ Coefficient of Determination and Root Mean Squared Error (RMSE) between Experimental Activity and Predicted Activities.}
\label{table:performance_metrics}
\end{table}

\begin{figure}[h!]
\centering
\includegraphics[width=1\linewidth]{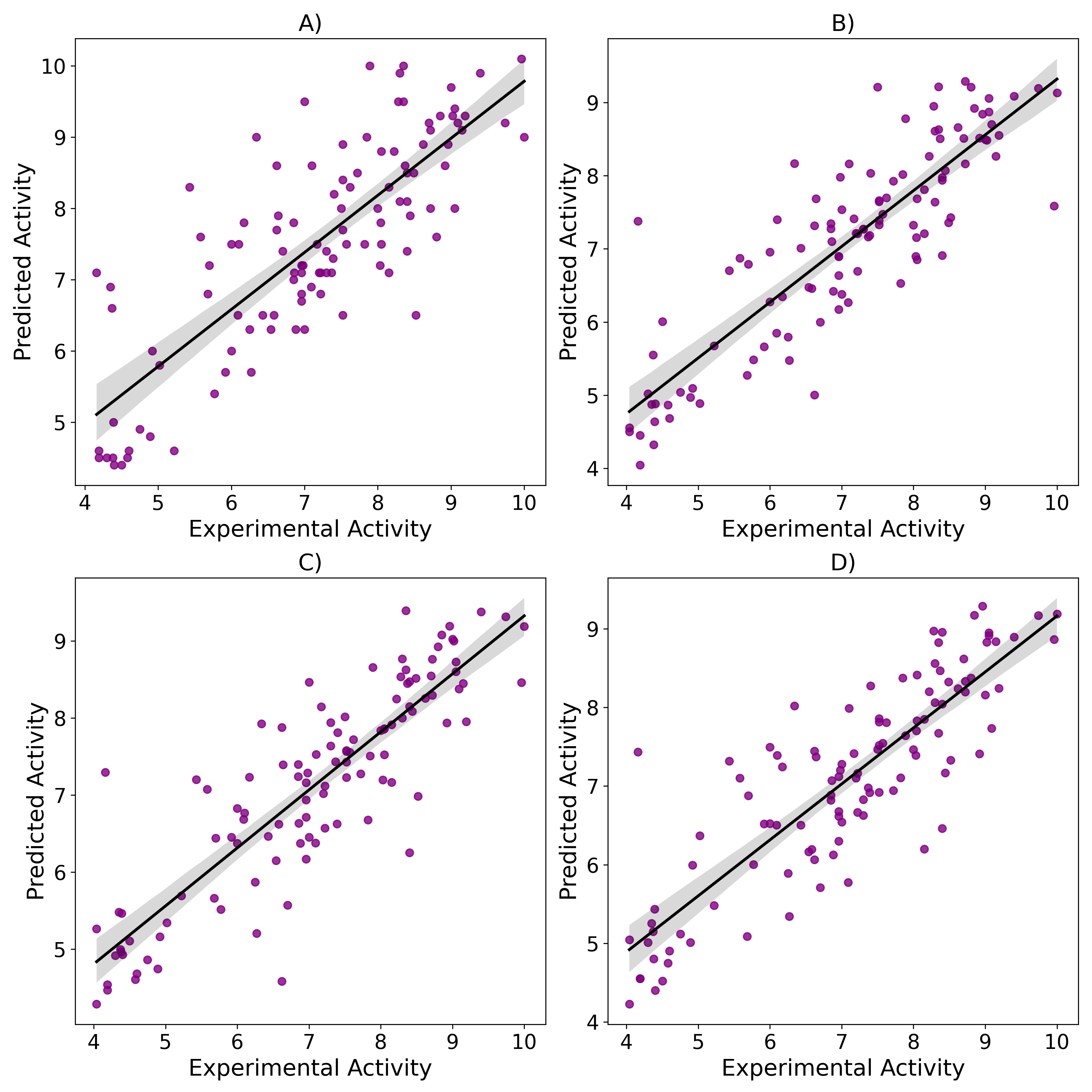}
\caption{Scatter plots representing the relationship between predicted and experimental activities of a validation dataset from cross validation folds for MMP8 protein target. (A) shows the activities predicted by the LLM model, while (B), (C), and (D) depict the results of CatBoost regression models trained on diverse input features; circular fingerprints with a radius of 3 and a 2048-bit vector size, RDKit descriptors, and Mol2Vec features, respectively. }
\label{fig:model_ensemble_nonw}
\end{figure}

\subsection{Interactive molecular design}



\begin{figure}[ht!]
\centering
\includegraphics[width=1\linewidth]{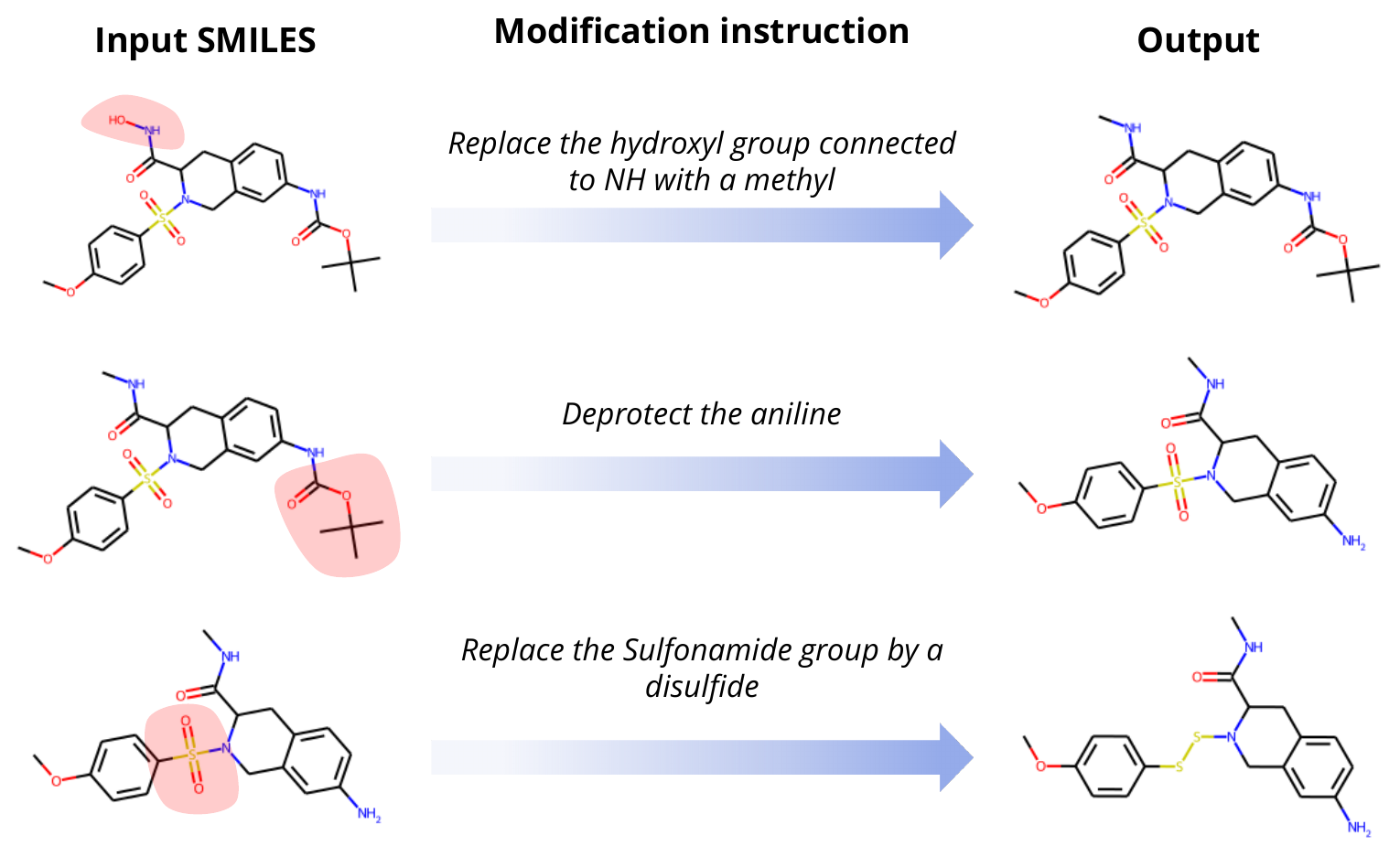} %
\caption{Overview of the iterative design process. Our tool aims to address the challenge of SMILES modification, which requires extensive understanding of structural chemistry and SMILES notations.}
\label{fig:iterative_design}
\end{figure}


We developed an interactive user interface that utilizes  LLMs to perform structural modifications. These modifications include adding, removing, or replacing functional groups, as well as more general physiochemical changes in steric hindrance, polarity, and hydrophobicity of different molecular fragments. Figure \ref{fig:iterative_design} shows some example input molecules, instructions, and the generated modified molecules. This tool not only facilitates the design and modification of desired molecules but also serves as an efficient tool for obtaining feedback from domain experts, which can be used in the context for further enhancement of molecular generation via ICL.








\section{Conflict of Interest}

This work was funded by Sanofi. All authors are Sanofi employees and may hold shares and/or stock options in the company.

\nocite{langley00}

\bibliography{main}
\bibliographystyle{icml2024}

\newpage
\appendix
\onecolumn
\section{Appendix}
\subsection{Technical details}\label{technical_details}


Our workflow utilizes the LiteLLM package, which allows for seamless usage of a wide variety of deployed LLMs through Bedrock, Hugging Face, OpenAI, etc., as the molecular generation engine. For interactive modification of generated molecules using expert instructions, we explored both the Claude 3 Sonnet and OpenAI GPT-4 models. Although further extensive benchmarks are needed, we empirically observed a deeper understanding of chemical structure and a better capability for following instructions for structural modification by GPT-4.

\begin{figure}[h!]
\centering
\includegraphics[width=0.7\linewidth]{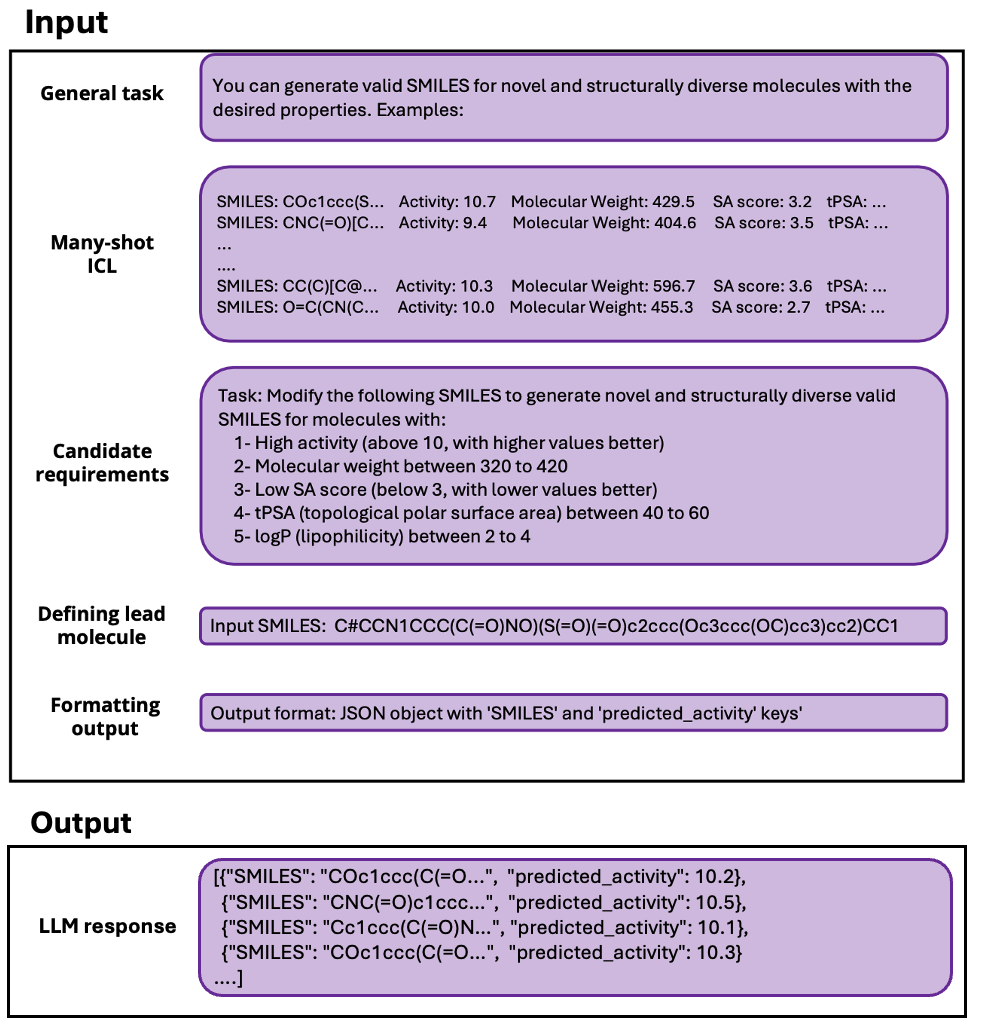}
\caption{Template used for the generation of new candidates through ICL. By providing clear instructions and a large and diverse pool of examples, the model learns underlying structural patterns that lead to the molecular property value. We need to explicitly indicate every detail around the task that was to be conducted. This includes the property requirements, the starting molecule, and how the output should be formatted.
By learning on the examples provided in the prompt, the model outputs new unseen candidates which should match the established criteria.}
\label{fig:icl_workflow}
\end{figure}

\begin{figure}[h!]
\centering
\includegraphics[width=0.7\linewidth]{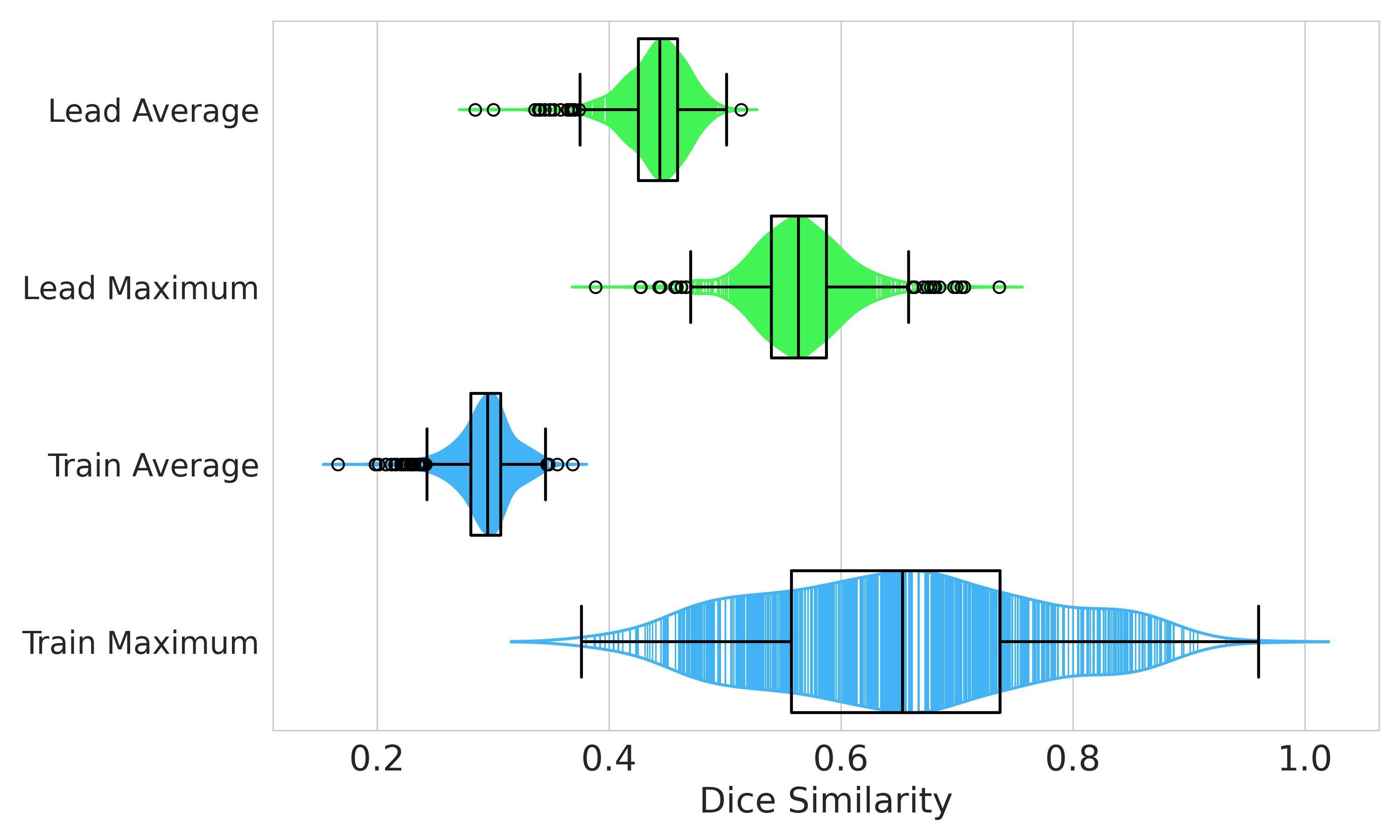}
\caption{Distribution of Average and Maximum Dice Similarity between Generated Molecules in the 500-Shot Setup and the Molecules in both Training and Lead Datasets.}
\label{fig:icl_workflow}
\end{figure}

\begin{figure}[h!]
\centering
\includegraphics[width=0.5\linewidth]{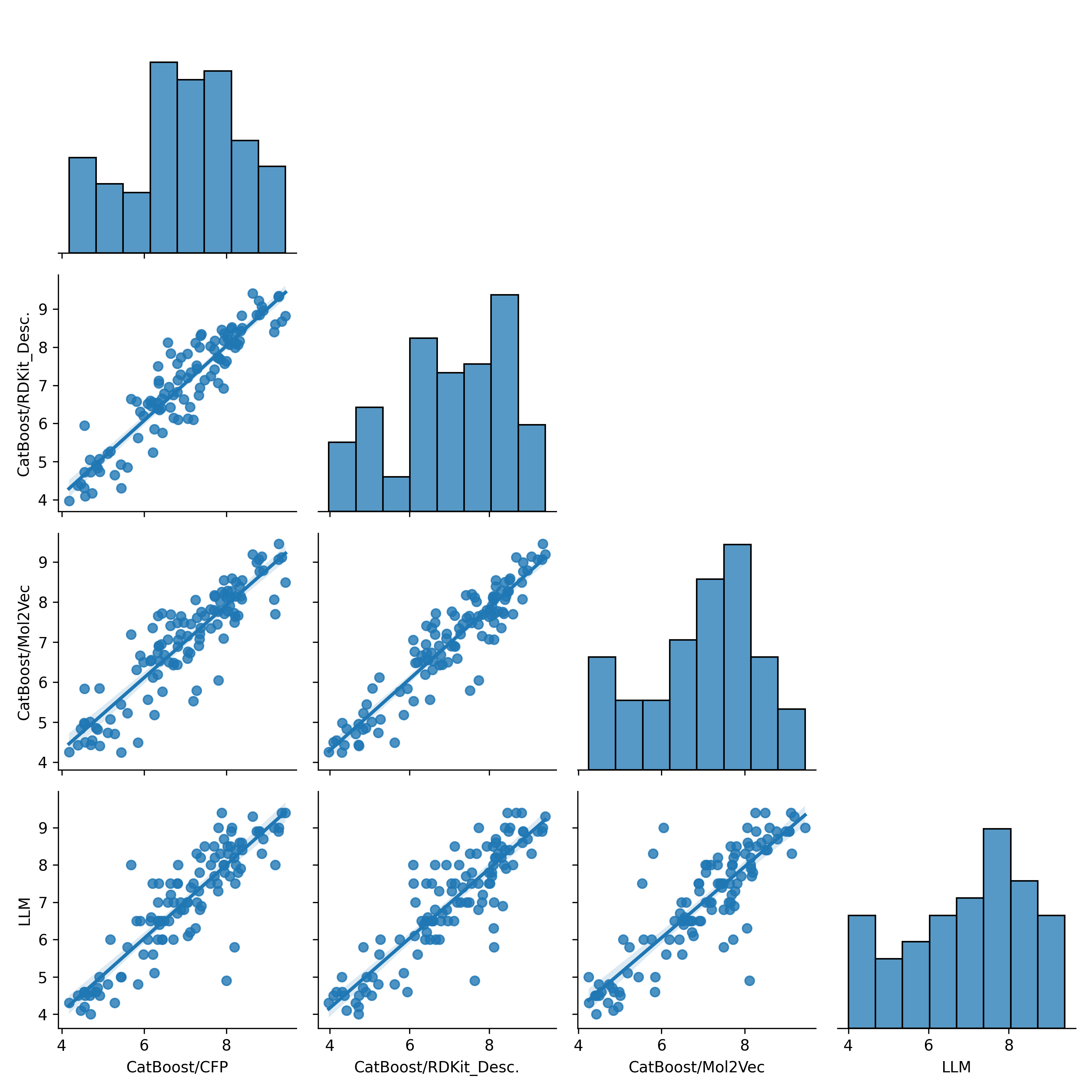}
\caption{Pairwise correlation plots of predicted activities from CatBoost models trained with different input features and LLM predictions on validation dataset}
\label{fig:pairplot}
\end{figure}

\subsection{Comparison to other SOTA molecular design tools }\label{sec:comparison_sota}

We compared the performance of our framework with REINVENT 4 which is a modern open-source generative AI framework for the molecular design. Reinvent4 uses a reinforcement learning for molecular design using a SMILES as the representation of the molecules. The algorithm for molecular optimization implemented in the package utilizes Mol2Mol - a transformer-based conditional prior model. It was trained by systematic and exhaustive exploration of chemical space while being regularized on the molecular (Tanimoto) similarity. Therefore, the similarity can be directly to the corresponding negative log-likelihood of the generated molecular representation.
In coarse of molecular optimization, the likelihood of generated molecule is additionally biased by a scalar score evaluating the targeted chemical properties as well as diversity filter that promotes diversity of the generated molecules by penalization of already explored SMILES and scaffolds.
 
The molecular optimization pipeline was run separately for all lead structures (pool50 dataset) {for each protein target}. The reinforced learning (“staged learning” run type) starting from a single lead molecule was conducted for 300 epoch for a batch of 3 molecules.  The following properties was calculated within Reinvent package and used for calculation of the score – molecular weight, partition coefficient (ClogP), synthetic accessibility (SA) score and topological polar surface area (tPSA). In addition, the prediction of activity was calculated utilizing a custom reinvent plugin for a catboost compound activity model. Each property was normalized to the range [0,1] by application of one-sided or two-sided sigmoid functions. The final score was evaluated as the geometric mean of individual property scores. The diversity filter on Murcko scaffolds was applied with buckets for 10 molecules and the additional molecules with the same scaffold were scored zero. The detailed parameters of the runs and property transformations are listed in a reinvent configuration file.

\begin{figure}[h!]
\centering
\includegraphics[width=0.9\linewidth]{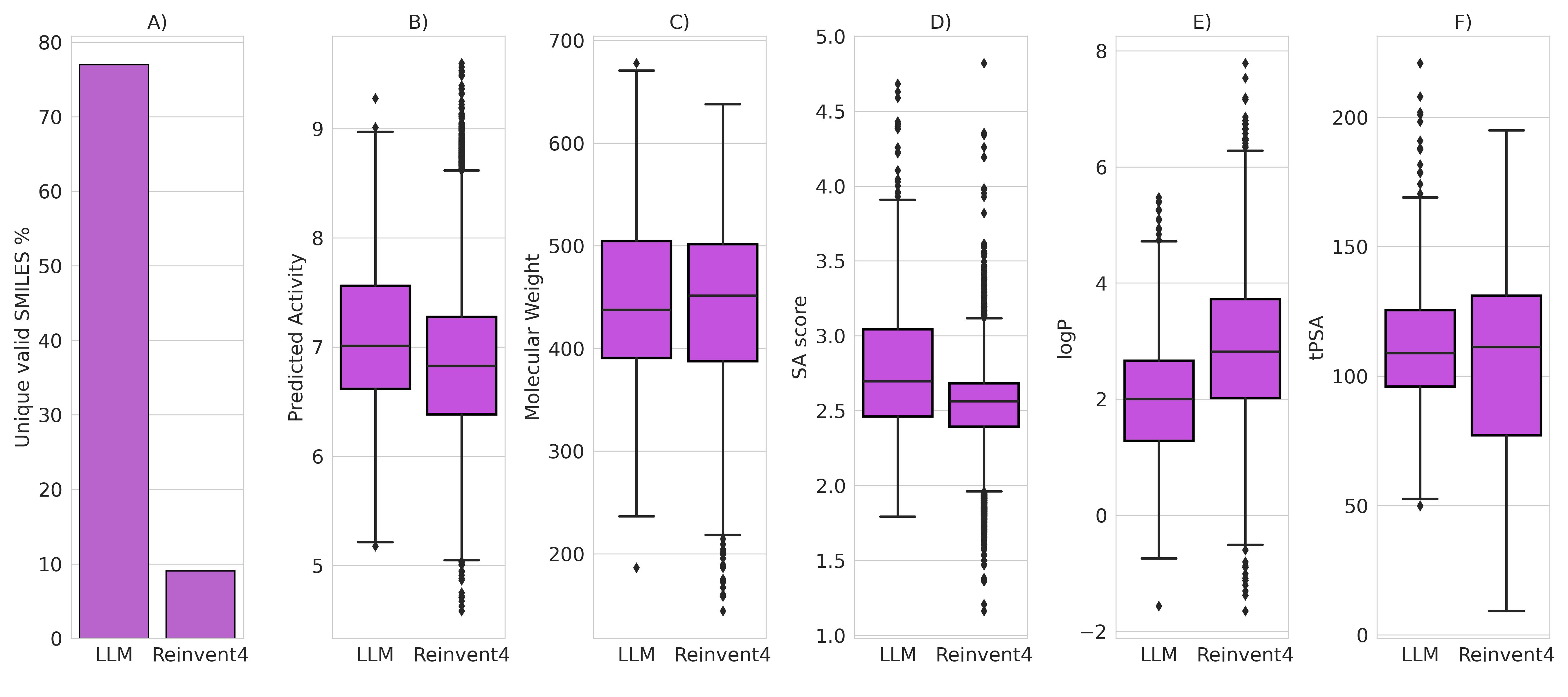} %
\caption{Performance comparison of generated molecules through our LLM framework and Reinvent4 using different metrics on MMP8 protein target. A) Rate of unique, new, and valid generated SMILES. B) Predicted activity. C) Molecular weight. D) Synthetic Accessibility (SA) score. E) LogP. F) Topological Polar Surface Area (tPSA).}
\label{fig:comparison_sota}
\end{figure}

\begin{figure}[h!]
\centering
\includegraphics[width=0.9\linewidth]{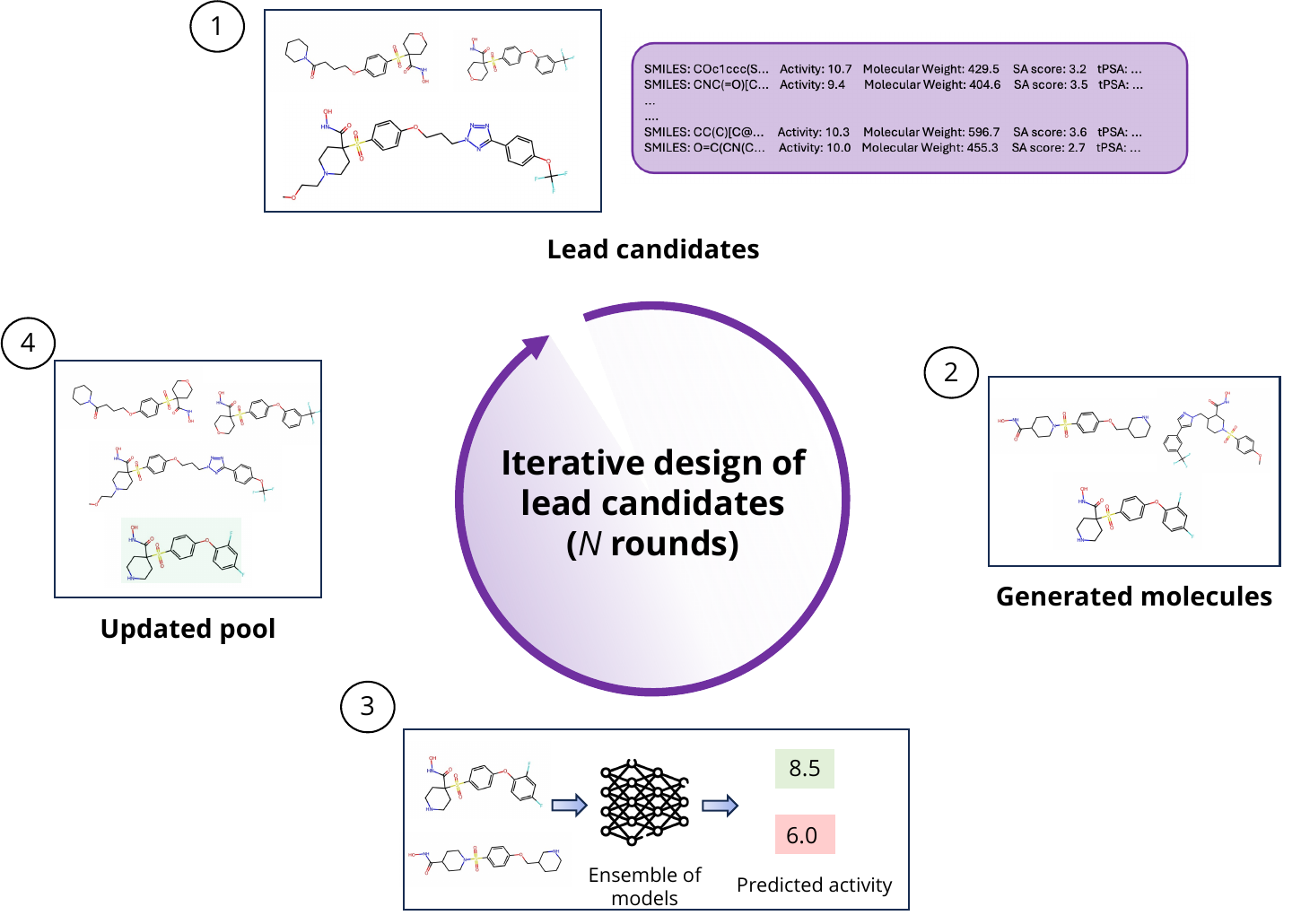} %
\caption{Lead optimization iterative process using LLMs. Process starts with a pool of experimental lead candidates that are inputted into the LLM through the established format together with other instructions. When molecules are generated, the ensemble of models validate the property values. Candidates with values above 0.8 percentile are included in the pool and used in next iteration.}
\label{fig:lead_optimization}
\end{figure}


\end{document}